\documentclass[lettersize,journal]{IEEEtran}
\usepackage{amsmath,amsfonts}
\DeclareUnicodeCharacter{FF0C}{,}
\usepackage{algorithmic}
\usepackage{algorithm}
\usepackage{array}
\usepackage[caption=false,font=normalsize,labelfont=sf,textfont=sf]{subfig}
\usepackage{textcomp}
\usepackage{stfloats}
\usepackage{url}
\usepackage{verbatim}
\usepackage{graphicx}
\usepackage[absolute,overlay]{textpos} 
\usepackage{amsmath} 
\graphicspath{ {image/} }  
\usepackage{float} 
\usepackage{stfloats} 
\usepackage{wrapfig} 
\usepackage{cite}
\usepackage{booktabs} 
\usepackage{tabularx} 
\usepackage{makecell} 
\usepackage{xcolor}

\usepackage{hyperref}
\usepackage{xcolor} 
\hypersetup{
  colorlinks=true,
  linkcolor=blue,
  filecolor=blue,  
  citecolor=blue,
  urlcolor=blue,
  pdftitle={Overleaf Example},
  pdfpagemode=FullScreen,
}
\usepackage{multirow} 
\DeclareUnicodeCharacter{221A}{$\sqrt{}$}
\usepackage{booktabs}
\usepackage{orcidlink}
\usepackage{amsmath}
\usepackage[utf8]{inputenc}  
\usepackage{newunicodechar}
\newunicodechar{￥}{\yen}
\usepackage{adjustbox}

\title{WaveGuard: Robust Deepfake Detection and Source Tracing via Dual-Tree Complex Wavelet and Graph Neural Networks}

\author{
Ziyuan He\textsuperscript{},
Zhiqing Guo\textsuperscript{},~\IEEEmembership{Member,~IEEE},
Liejun Wang\textsuperscript{}, 
Gaobo Yang\textsuperscript{},\\
Yunfeng Diao\textsuperscript{},
and Dan Ma\textsuperscript{},~\IEEEmembership{Member,~IEEE}%

\thanks{Ziyuan He, Zhiqing Guo, Liejun Wang, and Dan Ma are with the College of Computer Science and Technology, Xinjiang University, Urumqi, 830017, China. Zhiqing Guo and Liejun Wang are also with the Xinjiang Multimodal Intelligent Processing and Information Security Engineering Technology Research Center, Urumqi, 830017, China (e-mail: 107552304059@stu.xju.edu.cn; \{guozhiqing,wljxju,madan\}@xju.edu.cn).}%
\thanks{Gaobo Yang is with College of Computer Science and Electronic Engineering, Hunan University, Changsha, 410082, China. (e-mail: yanggaobo@hnu.edu.cn).}%
\thanks{Yunfeng Diao is with School of Computer Science, Hefei University of Technology, Hefei, 230009, China (e-mail: diaoyunfeng@hfut.edu.cn).}%

}

\begin{document}

\maketitle
\renewcommand{\figurename}{Fig.}

\begin{abstract}
Deepfake technology has great potential in the field of media and entertainment, but it also brings serious risks, including privacy disclosure and identity fraud.
To counter these threats, proactive forensic methods have become a research hotspot by embedding invisible watermark signals to build active protection schemes.
However, existing methods are vulnerable to watermark destruction under malicious distortions, which leads to insufficient robustness.
Moreover, embedding strong signals may degrade image quality, making it challenging to balance robustness and imperceptibility.
Although watermarked images look natural, their underlying structures are often different from the original images, which is ignored by traditional watermarking methods.
To address these issues, this paper proposes a proactive watermarking framework called WaveGuard, which explores frequency domain embedding and graph-based structural consistency optimization.
In this framework, the watermark is embedded into the high-frequency sub-bands by dual-tree complex wavelet transform (DT-CWT) to enhance the robustness against distortions and deepfake forgeries.
By leveraging joint sub-band correlations and selected sub-band combinations, the framework enables robust source tracing and semi-robust deepfake detection.
To enhance imperceptibility, we propose a Structural Consistency Graph Neural Network (SC-GNN) that constructs graph representations of the original and watermarked images to ensure structural consistency and reduce perceptual artifacts.
Experimental results show that the proposed method performs exceptionally well in face swap and face replay tasks. 
The code has been published at https://github.com/vpsg-research/WaveGuard.
\end{abstract}


\section{Introduction}
\label{sec:introduction}
\begin{figure}[t] 
    \centering
    \includegraphics[width=\columnwidth, keepaspectratio]{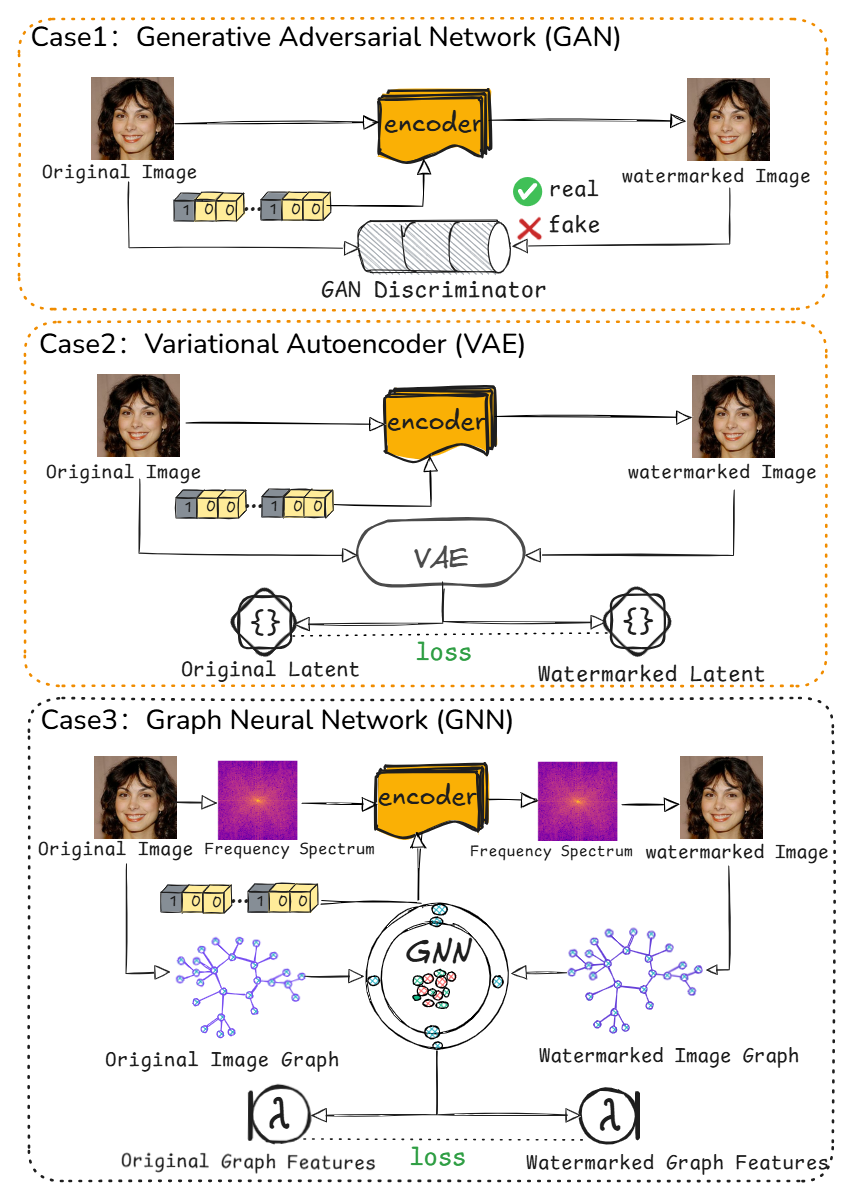} 
    \caption{Three categories of approaches for balancing robustness and invisibility in proactive deepfake detection.}
    \label{approaches}
\end{figure}

\IEEEPARstart{T}{he} rapid development of artificial intelligence generated content (AIGC) has greatly improved the ability of automation and innovation.
However, its wide application also brings risks, especially the deepfake technology, which can generate forged media, threaten privacy, encourage identity theft and spread false information\cite{wang2024deepfake}. 
Deepfake generation techniques mainly includes face swap and face replay.
Face swap\cite{swap1}, \cite{fsrt}, \cite{cscs}, \cite{uniface}, \cite{hififace}, replaces the identity features of the target into the source image while preserving the background and semantic information.
Face replay\cite{replace3,replace8} manipulates expressions or poses to map dynamic features onto the target image while maintaining identity consistency.
Although deepfake has positive application in some fields, its abuse has caused serious social concern.
Thus, it is very important to develop effective detection methods.

Currently, most deepfake detection methods rely on passive detection, which determines the authenticity of images by analyzing anomalous features present in forged content.
Typically, quantization-based three-dimensional convolutional neural networks (3D CNNs) are employed to enhance the discrimination of spatiotemporal features \cite{3DCNNs}. Other studies use perturbation-domain alignment to improve cross-domain robustness \cite{robust}. In addition, researchers have explored frequency-domain analysis to detect spectral distortions and encoding residuals \cite{Frequency}.
Lightweight video feature modeling methods have also been proposed to boost detection efficiency while maintaining high accuracy \cite{LED}.
Although these methods are effective in content authenticity verification, it usually comes into play after the generation and dissemination of forged content, and lacks effective traceability and interpretability.

Recently, the concept of proactive approach has emerged as a preventive strategy, and its core idea is to embed invisible signals when the original content is released.
This approach is mainly divided into proactive distortions\cite{RUiz}, \cite{cmua}, \cite{TAFIM}, \cite{anti}, \cite{df-rap} and proactive watermarks\cite{artificial01}, \cite{artificial02}, \cite{facesigns}, \cite{faceguard}, \cite{sepmark}, \cite{wu}, \cite{wang}, \cite{lampmark}, \cite{Dualdefense}. The former interferes with the generation process of a deepfake model by embedding invisible adversarial perturbation in the original image, destroying or counteracting the effect of the deepfake model, and making the generated content appear visually abnormal.
While this provides early protection, it lacks traceability and verification capabilities. 
In addition, proactive distortions may also weaken the usability of deepfake technology in legitimate applications.
In contrast, the proactive watermarks method is to embed the watermark information in the original content in advance, and then realize the dual functions of traceability and detectability by detecting the absence or change of the watermark.
This method ensures copyright protection and source tracing, especially avoids the unexplained problems in the passive deepfake detection results.
However, the existing proactive watermarks methods face challenges in robustness, particularly when dealing with complex deepfake attacks. 
As deepfake technology advances, the forgery techniques have become more subtle and sophisticated, making it increasingly difficult to extract watermarks effectively.
Meanwhile, maintaining invisibility is equally important.
Improving robustness often requires stronger embedding, which may introduce perceptual artifacts.
Conversely, enhancing invisibility will reduce the robustness against distortion.
Thus, balancing robustness and invisibility has become a key challenge for proactive watermarks.

To better understand existing approaches for balancing robustness and invisibility, Fig. \ref{approaches} presents three different categories of deep learning-based watermarking frameworks.
The first two cases adopt spatial domain embedding strategy, and hide watermark information by directly modifying image pixel values.
Although the spatial domain method is simple and efficient, it still faces significant challenges when confronted with complex attacks.
Because pixel modification inevitably damages the embedded watermark, this method makes the spatial watermark highly sensitive to distortion and deepfake attack.
In Case1, Generative Adversarial Network (GAN)\cite{GAN} generate high-quality watermarked images through adversarial training, which makes the watermark image visually similar to the original image to enhance invisibility.
However, GANs often face problems of unstable training and mode collapse, which result in inconsistent and unstable outputs.
In Case2, Variational Autoencoder (VAE)\cite{VAE} extract latent features from both the original and watermarked images and optimize their similarity through a loss function, aiming to enhance the watermarked image quality while preserving natural image structures.
However, due to the constraints of latent space, the reconstructed images generated by VAEs are often blurred and the detailed information is easily lost. Especially when different datasets are pretrained, the effect of feature extraction is not satisfactory.\par 
To address the limitations of existing methods in both robustness and invisibility, we propose a dual-channel watermarking framework called WaveGuard.
It integrates frequency-domain embedding with a graph neural network-based structural consistency constraint to support robust source tracing and deepfake detection.
Aiming at the problem that the local details are often lost and the directional texture disturbance often leads to the invalidation of the watermark, this framework innovatively uses the translation invariance and directional selectivity of the dual-tree complex wavelet transform (DT-CWT)\cite{DTCWT} to embed the watermark into the high-frequency sub-band, thus significantly enhancing its robustness to common distortions and Deepfake forgery.
In addition, the traceability and detection function of the watermark are realized by using the correlation between the joint sub-bands. By using different sub-band combinations, ensuring robustness for traceability and semi-robustness for detection.
To further enhance performance, the framework adopts dense connection strategies to promote efficient feature reuse and improve the robustness of the watermark.
Additionally, in order to solve the problem that traditional watermarking methods introduce obvious visual artifacts, this paper designs a structural consistency graph neural network (SC-GNN), as shown in Case 3 in Fig. \ref{approaches}.
Unlike VAE, which extracts the latent features of images, SC-GNN captures the spatial correlation of original images and watermarked images by constructing graphical representations.
The graph convolution network (GCN) further extracts and optimizes the structural features, which ensures that the watermark image maintains the integrity of the original image and reduces the perceptual differences.
Our contributions can be summarized as follows:\\
• We adopt a frequency-domain watermarking strategy using DT-CWT, leveraging high-frequency sub-bands to enhance resistance against distortions and forgeries.
By utilizing joint sub-band correlations for robust and semi-robust embedding,
the method ensures reliable source tracing and deepfake detection.\\
• We propose a Structural Consistency Graph Neural Network (SC-GNN), which leverages the internal structural relationships of the image to enhance the imperceptibility of the watermarked image. This method constructs graph representations of both the original and watermarked images, capturing spatial correlations to better maintain visual consistency.\\
• We construct an end-to-end dense connection network WaveGuard, which improves the flexibility and resilience of watermark embedding by optimizing feature transmission and reuse. Experimental results validate the advantages of our method in watermark recovery and deepfake detection, achieving superior robustness and imperceptibility compared to existing approaches. 

The remainder of this paper is organized as follows. Section~\ref{sec:related_work} reviews the related work on proactive defense, digital watermarking, and proactive forensics. 
Section~\ref{sec:methodology} presents the proposed \textit{WaveGuard} framework in detail. Section~\ref{sec:experiments} describes the experimental setup and reports the evaluation results. Section~\ref{sec:conclusion} concludes the paper.

\section{RELATED WORKS}
\label{sec:related_work}
\subsection{Proactive Distortions}
Proactive defense strategy destroys or counteracts the generation effect of a deepfake model by embedding invisible adversarial perturbation in the original image. 
This method provides preventive protection before the image is released, ensuring that when the image is used in deepfake, the generated content will appear visually abnormal. 
For example, Ruiz et al.\cite{RUiz} proposed a confrontational attack method that destroyed the conditional image translation network through invisible perturbations to prevent deepfake face manipulation. 
Similarly,
CMUA\cite{cmua} significantly reduces the forgery ability of several deepfake models by generating antagonistic perturbations.
In addition, TAFIM\cite{TAFIM} uses an attention mechanism to fuse specific perturbations and generate perturbations that cannot be identified by the model, thereby preventing image manipulation. 
Furthermore, Anti-Forgery\cite{anti} interferes with the generation of deepfake content by introducing perceptual perturbations into the Lab color space, creating significant artifacts and maintaining robustness against input transformations. 
At the same time, DF-RAP\cite{df-rap} provides robust image protection by modeling the social network compression mechanism to defend against deepfake.
Although proactive defense can interfere with the generation of deepfake at an early stage, it still has some key limitations.
For example, powerful generation models may bypass or neutralize these perturbations, thus weakening their effectiveness.
In addition, proactive distortions lacks tracking ability and post-verification, and it is difficult to judge whether the image has been tampered with after publication.

\subsection{Proactive Watermarks}
Digital watermarking is a technology to protect copyrights, verify integrity, or source tracing by embedding hidden information in digital media.
HiDDeN\cite{hidden} is the first end-to-end trainable digital robust watermarking framework. 
It uses the encoder-decoder structure of deep neural networks to hide information and achieves efficient recovery under various image distortions, opening a new research direction in the field of digital watermarking. 
To address challenges introduced by JPEG compression, MBRS\cite{mbrs} introduced a robustness-oriented approach by dynamically switching among different encoding settings, including real and simulated JPEG as well as noise-free pathways, thereby enhancing adaptability.
Similarly, CIN\cite{cin} further advanced robustness and invisibility by combining a reversible transform with attention mechanisms. 
In addition, TSDL\cite{tsdl} replaced the conventional single-stage architecture with a sequential dual-phase model, resulting in more stable embedding and faster convergence during training.
Tang et al.\cite{tang2022highly}proposed a two-stage reversible watermarking framework that embeds watermarks into pseudo-Zernike moments using adaptive normalization and optimized DC-QIM, achieving strong resistance to distortions while enabling exact image recovery under benign conditions.
ARWGAN\cite{arwgan} introduces an attention-guided GAN-based watermarking model that enhances robustness through adversarial training under diverse distortions.
Similarly, Huang et al.\cite{huang2023robust} proposed a texture-aware watermarking method that embeds in high-texture regions using an SSIM-guided regression model, achieving robustness to various attacks while maintaining perceptual quality.
In addition, Wang et al.\cite{wang2024robust} proposed a watermarking framework using DCT-SVD embedding and a template-enhanced network to achieve robust synchronization and accurate extraction under geometric distortions.
More recently, Zhu et al.\cite{zhu2024lite} proposed a deep watermarking framework combining Lite-HRNet and a dual U-Net encoder, achieving robustness to screen-shooting and printing while preserving imperceptibility via adaptive noise and adversarial training.
For cross-media distortion (such as print scanning, print camera and screen camera, etc.), researchers have proposed various solutions, including LFM\cite{lfm}, Stegstamp \cite{stegastamp}, RIHOOP \cite{rihoop}, PIMoG \cite{pimog}, GDS-WM \cite{GDS-WM}, and WRAP\cite{WPAP}.
These methods target different distortion types by designing optimized strategies to ensure stable watermark extraction and verification after physical printing and re-digitization.

In the emerging field of proactive face forensics, existing methods are limited.
Yu et al. \cite{artificial01, artificial02} proposed an artificial fingerprint embedded within generative models for deepfake detection and source tracing, demonstrating strong scalability and robustness against complex noise.
FaceSigns\cite{facesigns} differs from traditional robust watermarking by emphasizing robustness under common distortions while selectively failing under malicious distortions, thus achieving more accurate deepfake detection. 
Similarly, FaceGuard\cite{faceguard} achieves proactive deepfake detection by embedding semi-robust watermarks.
Building on these efforts, Sepmark\cite{sepmark} proposed a separate watermarking framework that effectively addresses both common and malicious distortions by combining robust and semi-robust watermarking using a single encoder and separable decoder.
At the same time, Wu et al.\cite{wu} proposed a new framework that adjusts robust watermarks into adversarial ones to improve detection and tracing capabilities. It also supports passive detection, making it easier to combine proactive and passive detection.
Wang et al. \cite{wang} proposed an identity-perceptual watermarking framework that enables unified deepfake detection and source tracing by embedding content-dependent semantic watermarks.
EditGuard\cite{editguard} proposes a proactive watermarking framework that enables zero-shot tamper localization and copyright protection via dual image-bit steganography.
In addition, Lampmark\cite{lampmark} uses an untrained pre-defined pattern watermark to actively defend against deepfake, combining structural sensitivity and robustness to achieve efficient detection.
With the growing use of multifunctional watermarking, Dual Defense\cite{Dualdefense} employs an original-domain feature impersonation strategy to jointly embed adversarial and traceable watermarks, thereby actively disrupting face swap generation and enabling reliable traceability.

\begin{figure}[t] 
    \centering 
    \includegraphics[scale=0.5]{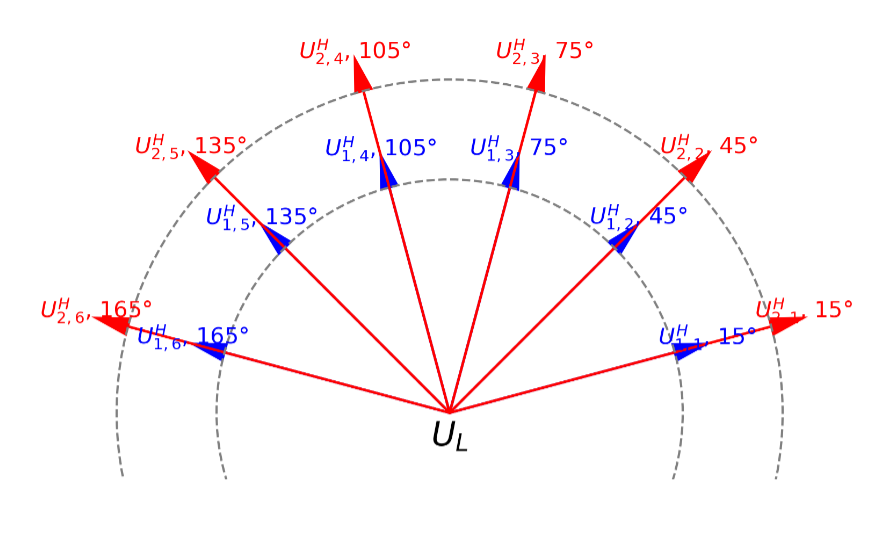} 
    \caption{Two-level dual-tree complex wavelet transform decomposition.} 
    \label{FIG:2}
\end{figure}

Overall, although the existing watermarking methods show some robustness under normal distortion conditions, they often fail in the face of advanced deepfake operations.
This kind of attack will reconstruct the semantic content and structural layout of the image, and the deepfake model usually re-renders the face region in a generative way. 
However, most watermarking methods only embed low-level signals in pixel space or traditional frequency domain, which lacks the adaptability to semantic content changes.
Meanwhile, the deepfake model has the ability of end-to-end optimization, which can automatically suppress or erase the imperceptible disturbance signal during the generation process, resulting in the watermark information being overwritten.
In this paper, WaveGuard embeds watermarks into directional frequency sub-bands and incorporates graph neural networks to reinforce structural consistency. Although it does not explicitly model semantic content, this structure-aware design exhibits strong resilience to distortions induced by semantic-level manipulations (e.g., generative face reconstruction), enabling reliable extraction while maintaining a favorable trade-off between robustness and invisibility.

\section{METHODOLOGY}
\label{sec:methodology}
\subsection{Preliminaries}
\subsubsection{Dual-tree complex wavelet transform}
Dual-Tree Complex Wavelet Transform (DT-CWT) is a wavelet transform with translation invariance and limited redundancy and its redundancy factor is 2. 
Unlike traditional DWT, DT-CWT decomposes the image into six directional high-frequency sub-bands, corresponding to angles of $15^\circ$, $45^\circ$, $75^\circ$, $105^\circ$, $135^\circ$, and $165^\circ$. 
This allows it to accurately capture the directional features and details of the image.
As shown in Fig. \ref{FIG:2}, DT-CWT adopts a circular layout to illustrate the directional and hierarchical structure of two-level decomposition. 
The first-level decomposition splits the image into six high-frequency sub-bands ($U^H_{1,d}$, where $d$ takes values from 1 to 6) and one low-frequency sub-band ($U^L$ at the center). 
The second-level decomposition further splits the first-level low-frequency sub-band, generating six new high-frequency sub-bands ($U^H_{2,d}$) and one low-frequency sub-band, with the size of each sub-band halved.
Blue arrows represent the first-level high-frequency sub-bands, and red arrows represent the second-level high-frequency sub-bands, with consistent directions across both levels. 
DT-CWT possesses multi-scale and multi-directional analysis capabilities, which precisely align with the need for accurate characterization of texture boundaries and directional details in watermarking.
In the face of compression, blurring, and sophisticated deepfake forgeries, its approximate translation invariance offers a stable and reliable embedding space in the frequency domain, thereby enhancing the robustness of watermark signals against fine-grained distortions. 
 
\subsubsection{Joint Sub-bands}

Joint sub-band correlation was first proposed by Huan et al.\cite{Exploring}, who significantly enhanced the robustness of video watermarking through the directional selectivity and translation invariance of DT-CWT. 
To further verify its role in image watermark embedding, we designed a simple experiment. 
\begin{figure} 
    \raggedright 
    \hspace*{-0.5cm} 
    \includegraphics[scale=0.16]{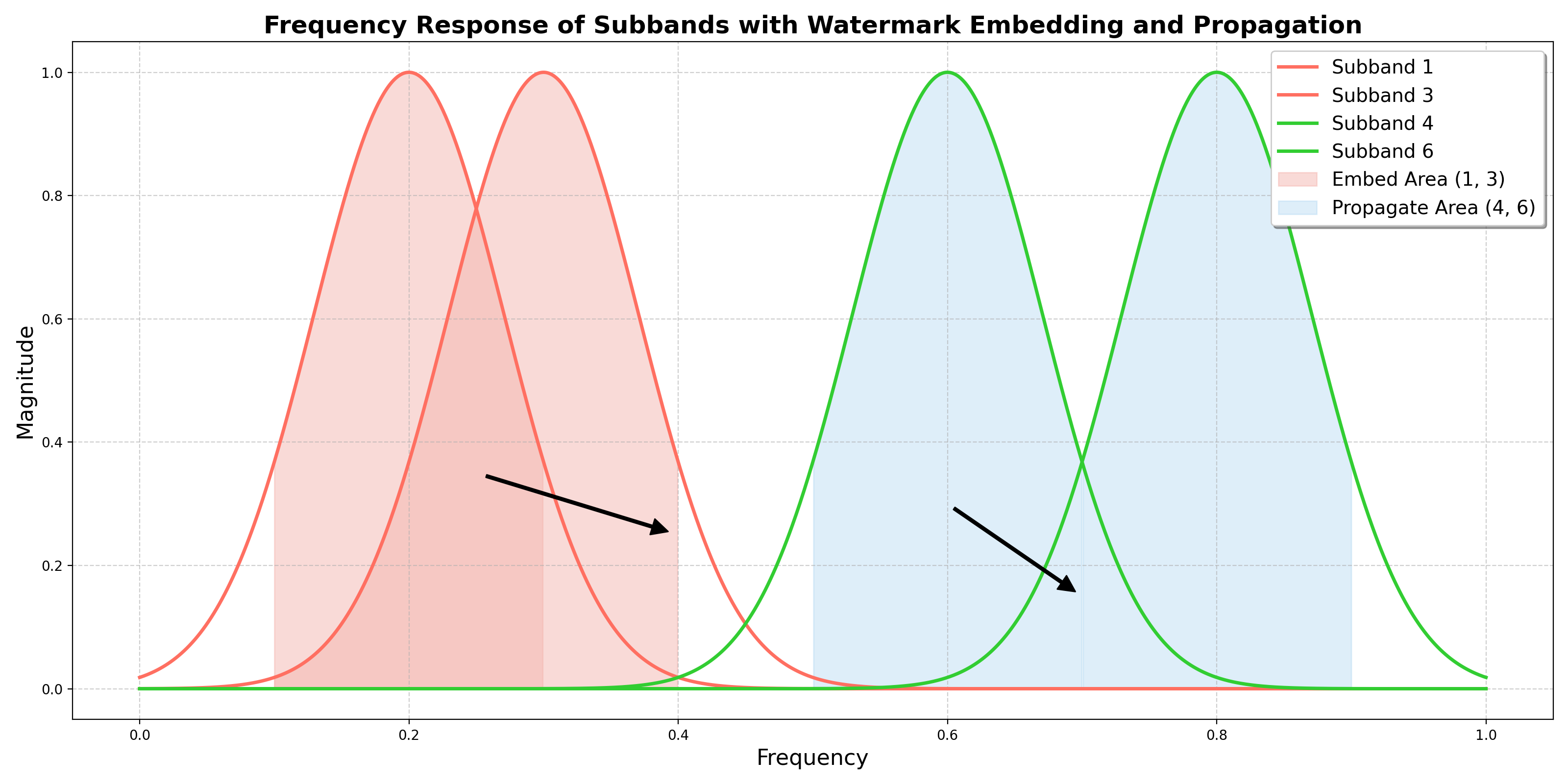} 
    \caption{Frequency response of four sub-bands with watermark embedding and propagation. Sub-band 1 and sub-band 3 (pink) represent the watermark embedding regions, while sub-band 4 and sub-band 6 (blue) indicate the propagation of watermark information due to sub-band redundancy.}
    \label{FIG:3}
    \vspace{-0.5cm}
\end{figure}
A standard grayscale image (256 × 256 pixels) was decomposed by DT-CWT to obtain a low-frequency sub-band and six high-frequency sub-bands. 
Watermarks were embedded in sub-band 1 and sub-band 3. The results show that, although the watermark is only embedded in sub-band 1 and sub-band 3, sub-band 4 and sub-band 6 carry some watermark information due to the redundancy between sub-bands. 
The correlation with the original watermark is significantly high.
Fig. \ref{FIG:3} illustrates the distribution of the watermark in the frequency domain: pink represents the watermark embedding area, blue indicates the propagation area, and arrows show the propagation path.
There is natural redundancy between the joint sub-bands, which enables the watermark information to spread in multiple high-frequency sub-bands, so it still has strong recovery ability under local tampering and frequency disturbance.
Based on the correlation design of joint sub-bands, watermark information can spread among multiple sub-bands, thus maintaining stable extraction in different attack scenarios, which has become the key supporting mechanism for the system to achieve robust traceability and semi-robust deepfake detection.

\subsection{Model Architecture}

In order to realize a robust watermark embedding and detection system for deepfake, this paper designs a framework, which mainly consists of the following six parts:
(1) \textbf{\textit{Frequency Subband Selection Module (FSSM)}}: The frequency subband selection module converts the input image from the spatial domain to the frequency domain and extracts specific frequency sub-band information.
(2) \textbf{\textit{Attention Mechanisms}}: To enhance watermark imperceptibility, attention mechanisms are introduced to focus on regions more suitable for embedding. By highlighting visually subtle yet semantically important areas, the framework improves watermark invisibility while preserving robustness against distortions.
(3) \textbf{\textit{Encoder (En)}}: The encoder uses a deep learning network to fuse the specific sub-band information output by the frequency subband selection module  with the watermark information, generating the embedded image.
(4) \textbf{\textit{Noise Pool Layer (NPL)}}: Simulates common distortions and malicious distortions in actual scenarios to generate the distorted image.
(5) \textbf{\textit{Separable Decoder}}: Includes DensTracer and DensDetector, which are used to extract watermarks under distorted conditions to realize image traceability and detect deepfake, respectively.
(6) \textbf{\textit{Structural Consistency
Graph Neural Network (SC-GNN)}}: By comparing the image characteristics before and after watermark embedding, the watermark embedding process is optimized to ensure the visual consistency of the watermark image and improve the invisibility of the watermark.

\begin{figure*}[t] 
    \centering
    \includegraphics[width=0.97\textwidth]{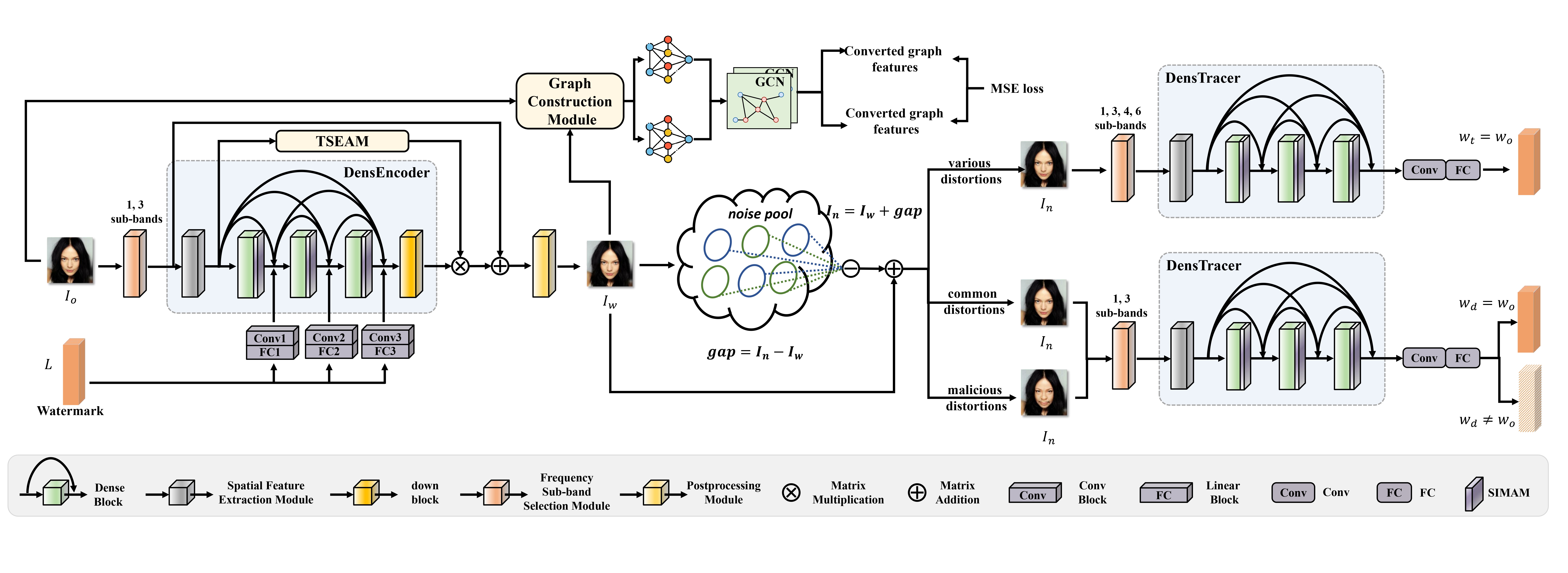} 
    \caption{
        Overall architecture of WaveGuard.
        \textbf{F}requency \textbf{S}ubband \textbf{S}election \textbf{M}odule ($FSSM$) extracts and processes high-frequency sub-bands in the frequency domain, providing essential features for watermark embedding, tracing, and detection.
       \textbf{A}ttention \textbf{M}odule ($AM$) highlights salient regions in the image and guides the features to focus on key areas, thereby enhancing the robustness and effectiveness of watermark embedding.
        The encoder ($En$) embeds the watermark into the high-frequency sub-bands and reconstructs the watermarked image $I_w$. The \textbf{N}oise \textbf{P}ool \textbf{L}ayer ($NPL$) applies various distortions to $I_w$ to generate distorted images $I_n$. 
        The tracer decoder ($Tr$) extracts the watermark $w_t$ from $I_n$ to evaluate its robustness, while the decoder ($De$) extracts the correct watermark $w_d$ under common distortions and outputs random-like messages under malicious distortions. In addition, \textbf{S}tructural \textbf{C}onsistency \textbf{G}raph \textbf{N}eural \textbf{N}etwork ($SC-GNN$) extracts the structural features of the original image $I_o$ and the watermarked image $I_w$ by constructing graph representation to ensure the consistency between them.
    } 
    \label{fig:fullwidth_image}
\end{figure*}

\subsubsection{\textbf{\textit{Frequency Subband Selection Module (FSSM)}}}
The input image is first decomposed into three channels: Y, U, and V, with the U channel as the main processing object. 
Secondly, the U channel is decomposed in the frequency domain using the dual-tree complex wavelet transform (DT-CWT), obtaining a low-frequency sub-band (low\_pass) and high-frequency sub-bands (high\_pass). 
High-frequency sub-bands contain frequency information in six directions. By selecting sub-bands 1 and 3, key high-frequency details are extracted for watermark embedding, while sub-bands 4 and 6 are reserved for the subsequent traceability stage.
This module completes the frequency domain conversion and sub-band selection, providing input for watermark embedding and extraction to the encoder and two separable decoders.

\subsubsection{\textbf{\textit{Attention Mechanisms (AM)}}}
To enhance the imperceptibility of watermark embedding, we introduce two complementary attention mechanisms into the framework. The Tanh-based Spatial Embedding Attention Module (TSEAM) is an improved version of the classic Convolutional Block Attention Module\cite{cbam}, where the ReLU activation function is replaced with Tanh to better accommodate continuous feature representations within the [0,1] range, thereby preserving the continuity and integrity of the watermark. TSEAM operates across the entire network to highlight semantically important yet visually inconspicuous regions, guiding the global embedding strategy. The Spatial-wise Image Attention Module (SIMAM)\cite{simam} is applied after each DenseBlock to assign spatial weights to local feature maps, enhancing the model’s sensitivity to fine-grained regions and further improving the stability and robustness of watermark embedding.
\subsubsection{\textbf{\textit{Encoder (En)}}}
The high-frequency sub-band feature output extracted by the frequency subband selection modul undergo Conv-BN-ReLU processing in the Spatial Feature Extraction Module, generating a refined feature map that captures both spatial and fine-grained details. 
The watermark is transformed into a two-dimensional matrix via a linear layer, resized to match the feature map dimensions, and processed by a convolutional block to extract watermark-specific features. These features are then fused with the shallow spatial features.
A three-stage dense connection structure is used for hierarchical watermark embedding. At each stage, the feature map and watermark are concatenated and processed through the DenseBlock Module, which consists of a 1×1 convolution for dimensionality reduction and a 3×3 convolution for feature enhancement.
To guide the embedding process, the TSEAM is applied globally across the encoder to highlight semantically important regions, while the SIMAM is applied after each DenseBlock to refine local spatial features.
The combination of TSEAM and SIMAM enhances embedding quality by jointly optimizing global semantics and local detail preservation.
After the three-stage fusion, the encoder applies spatial downsampling to compress the feature maps, reducing their dimensions while retaining critical information.
The final compressed features generate the watermark difference map (wm\_diff), which is added to the input image to complete the watermark embedding process.
Watermark embedding is achieved using the following formula:
\begin{equation}
\boldsymbol{U_{\mathit{embed}} = U + \alpha \cdot \mathit{M} \cdot \mathit{wm}_{\mathit{diff}}}
\label{eq:1}
\end{equation} 
In Formula (1), $U$ represents the original high-frequency sub-

\noindent band, $\alpha$ controls the watermark embedding strength, $\text{M}$ is the mask generated by the attention mechanism to weight the watermark embedding area, and $\text{wm}_{\text{diff}}$ is the generated watermark difference map. Through this process, the watermark information is embedded into the original high-frequency sub-band, forming a watermarked high-frequency sub-band $U_{\text{embed}}$. The post-processing module uses the inverse dual-tree complex wavelet transform to reconstruct the watermarked U-channel, merges it with the Y and V channels to form watermarked YUV images, and converts them to RGB images $I_w$, completing the embedding process.

\subsubsection{\textbf{\textit{Noise Pool Layer (NPL)}}}In practical applications, when images are uploaded or spread through social media, the quality is often degraded due to common processing operations such as compression and blurring or noise interference. To this end, we have established two operation sets: Common Pool and Malicious Pool, which are used to assist framework training. Among them, Common Pool contains common benign distortions in daily use, while Malicious Pool introduces deepfake model to improve the robustness of watermark in response to malicious attacks.

\subsubsection{\textbf{\textit{Separable Decoder}}}
The two decoders adopt a DenseNet structure similar to that of the encoder. The decoder takes the high-frequency sub-band embedded with the watermark as input, generates the initial feature map through the feature extraction module, and gradually fuses the decoded information through the multi-layer DenseBlock module. Finally, the high-dimensional features are compressed by the down-sampling module to generate a single-channel watermark difference map, which is adjusted to the target size through convolution and interpolation, and finally restored to one-dimensional watermark information.
Tracer and Detector share the same structure in design but differ in sub-band selection and target tasks. To enhance robustness, Tracer uses the 1st, 3rd, 4th, and 6th high-frequency sub-bands to extract features and handle various distortions, achieving watermark traceability with the goal of $W_t = W_o$. The Detector decoder only uses the 1st and 3rd sub-bands to extract features for the semi-robust requirement of forgery and tampering detection. Its target is $W_d = W_o$ for normal distortions and $W_d \neq W_o$ for malicious distortions. This difference in sub-band selection ensures that the two decoders meet the requirements for robust extraction and efficient detection, respectively.

\subsubsection{\textbf{\textit{Structural Consistency Graph Neural Network (SC-GNN))}}}
Although the watermarked image may appear visually natural, its underlying structural distribution often deviates from the original image. Traditional watermarking methods typically overlook structural consistency, leading to local artifacts or edge distortions that degrade perceptual quality.
To address this, we propose a Structural Consistency Graph Neural Network (SC-GNN). 

Specifically, the $U$ channels of both the original and watermarked images are extracted as single-channel inputs and converted into sparse graphs based on 4-neighborhood connectivity (top, bottom, left, and right), where each pixel is treated as a node with its normalized intensity as the node feature.
The constructed graph is fed into a two-layer Graph Convolutional Network (GCN): the first layer maps the node features to 16 dimensions with ReLU activation, while the second projects them to scalars. A global average pooling operation is then applied to obtain a graph-level embedding vector. This embedding vector is used to quantify the structural difference between the original and watermarked images and serves as the input to the structural consistency loss to preserve structural consistency.
\subsection{\textbf{\textit{Loss Functions}}}
To optimize our frequency-domain watermarking framework for embedding, tracing, and detection, we define loss functions to optimize the encoder-decoder framework. \( \theta \) represents trainable parameters, with \( \theta_{co} \) and \( \theta_{ma} \) denoting common and malicious distortions sampled from the Noise Pool Layer (NPL). The details are as follows:

\subsubsection*{• Embedding Loss}

The goal of the embedding loss is to ensure that the watermarked high-frequency sub-band $U_{\text{embedded}}$ is highly similar to the original $U$-channel. It is defined as:

\begin{equation}
\mathcal{L}_{En}=\|U-E_n(FSSM(I_o),W_o)\|_2^2=\|U-U_{embedded}\|_2^2
\label{eq:encoder_loss}
\end{equation}
where $FSSM(I_o)$ represents the original image processed by the frequency subband selection module, and $W_o$ is the embedded watermark.

\subsubsection*{• Tracing Loss}

The tracing loss ensures that the robust decoder $Tr$ can accurately extract the embedded watermark $W_t$ from the watermarked image $I_w$, even after various distortions. It is defined as:

\begin{equation}
\resizebox{0.445\textwidth}{!}{$
\mathcal{L}_{Tr}=\|W_o-Tr(\theta,NPL(\theta_{co}+\theta_{ma},I_w))\|_2^2=\|W_o-W_t\|_2^2
$}
\label{eq:tracer_loss}
\end{equation}
where $NPL$ is the noise processing layer that simulates distortions.

\subsubsection*{• Detection Loss}
The detection loss ensures that the semi-robust decoder $De$ behaves differently when facing common and malicious distortions:

The extracted watermark $W_d$ from the commonly distorted image $I_{no}$ should match the original watermark $W_o$:

\begin{equation}
\resizebox{0.445\textwidth}{!}{$
\mathcal{L}_{De1} = \| W_o - De(\theta, NPL(\theta_{co}, I_w)) \|_2^2 = \| W_o - W_{d} \|_2^2
$}
\label{eq:decoder_loss_1}
\end{equation}

For malicious distortions, the output $W_d$ of the decoder $De$ should approach random guessing (represented as 0):

\begin{equation}
\resizebox{0.445\textwidth}{!}{$
\mathcal{L}_{De2} = \| 0 - De(\theta, NPL(\theta_{ma}, I_w)) \|_2^2 = \| 0 - W_{d} \|_2^2
\label{eq:decoder_loss_2}
$}
\end{equation}

\subsubsection*{• SC-GNN Loss}
The SC-GNN loss optimizes the structural consistency between the original image and the watermarked image. It is defined as:  

\begin{equation}
\mathcal{L}_{\text{GNN}} = \bigl\| \text{GCN}(\text{GCM}(I_o)) - \text{GCN}(\text{GCM}(I_w)) \bigr\|_2^2 = \| f_o - f_w \|_2^2
\label{eq:gnn_loss}
\end{equation}
where $\text{GCM}(I_o)$ and $\text{GCM}(I_w)$ denote the graph representations constructed from the original and watermarked images, respectively, using the Graph Construction Module (GCM). These graphs are subsequently processed by the Graph Convolutional Network (GCN) to extract structural features, denoted as $f_o = \text{GCN}(\text{GCM}(I_o))$ and $f_w = \text{GCN}(\text{GCM}(I_w))$. The SC-GNN loss minimizes the mean squared error (MSE) between these feature representations to enforce structural consistency.

\begin{figure*}[t] 
    \centering
    \includegraphics[width=\textwidth]{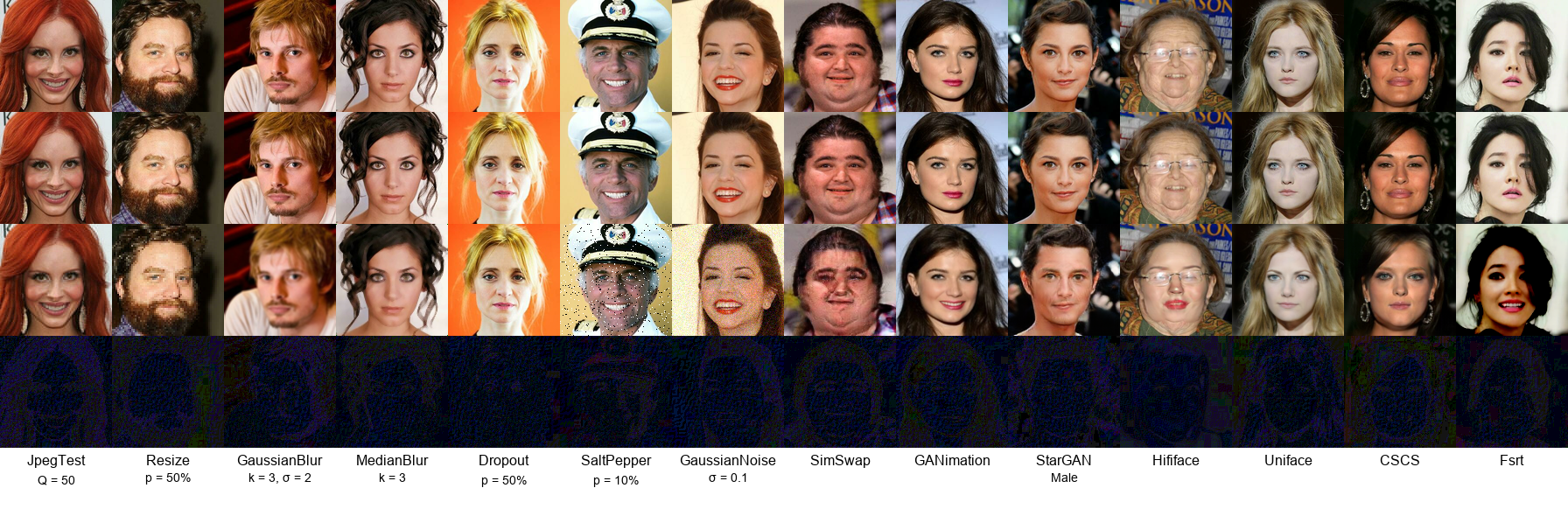} 
    \caption{Illustrates the visual effects of operations performed on watermarked images. The original image $I_o$ and the watermarked image $I_w$ are shown in the first and second rows, respectively. The results of applying different operations to the watermarked image, denoted as $I_n$, are displayed in the third row. The fourth row shows the watermarked residual signals, represented as $|I_o - I_w|$, which are processed to enhance the visualization of differences. All images are sized $128 \times 128$.}
    \label{fig:5}
\end{figure*}
\subsubsection*{• Total Loss}
The total loss is defined as a weighted combination of the above loss components, balancing their contributions during training. It is expressed as:

\begin{equation}
\mathcal{L}_{Total} = \lambda_1 \mathcal{L}_{GNN} + \lambda_2 \mathcal{L}_{En} + \lambda_3 \mathcal{L}_{Tr} + \lambda_4 \mathcal{L}_{De1} + \lambda_5 \mathcal{L}_{De2}
\label{eq:total_loss}
\end{equation}
where $\lambda_1, \lambda_2, \lambda_3, \lambda_4, \lambda_5$ are hyperparameters that control the weights of each loss component.

\section{EXPERIMENTS}
\label{sec:experiments}
\subsection{Implementation Details}
\textbf{Datasets.} This research is primarily based on the high-quality CelebA-HQ\cite{celeb1,celeb2}dataset, which contains 30,000 facial images and 6,217 unique identities. According to the official partitioning, it is divided into training, validation, and testing sets, where the training set contains 24,183 images, and the validation and test sets contain 2,993 and 2,824 images, respectively.
In addition, to evaluate the generalization performance on out-of-distribution data, we further test the model on the LFW\cite{LFW_tcsvt} dataset, which contains 13,233 unconstrained facial images from 5,749 identities.
To verify the adaptability and robustness of the algorithm across different resolutions and simulate the diverse requirements of practical application scenarios, the experiment uses two types of images with resolutions of 128×128 and 256×256 for testing.

\begin{figure*}[t]
    \centering
    \includegraphics[width=\textwidth]{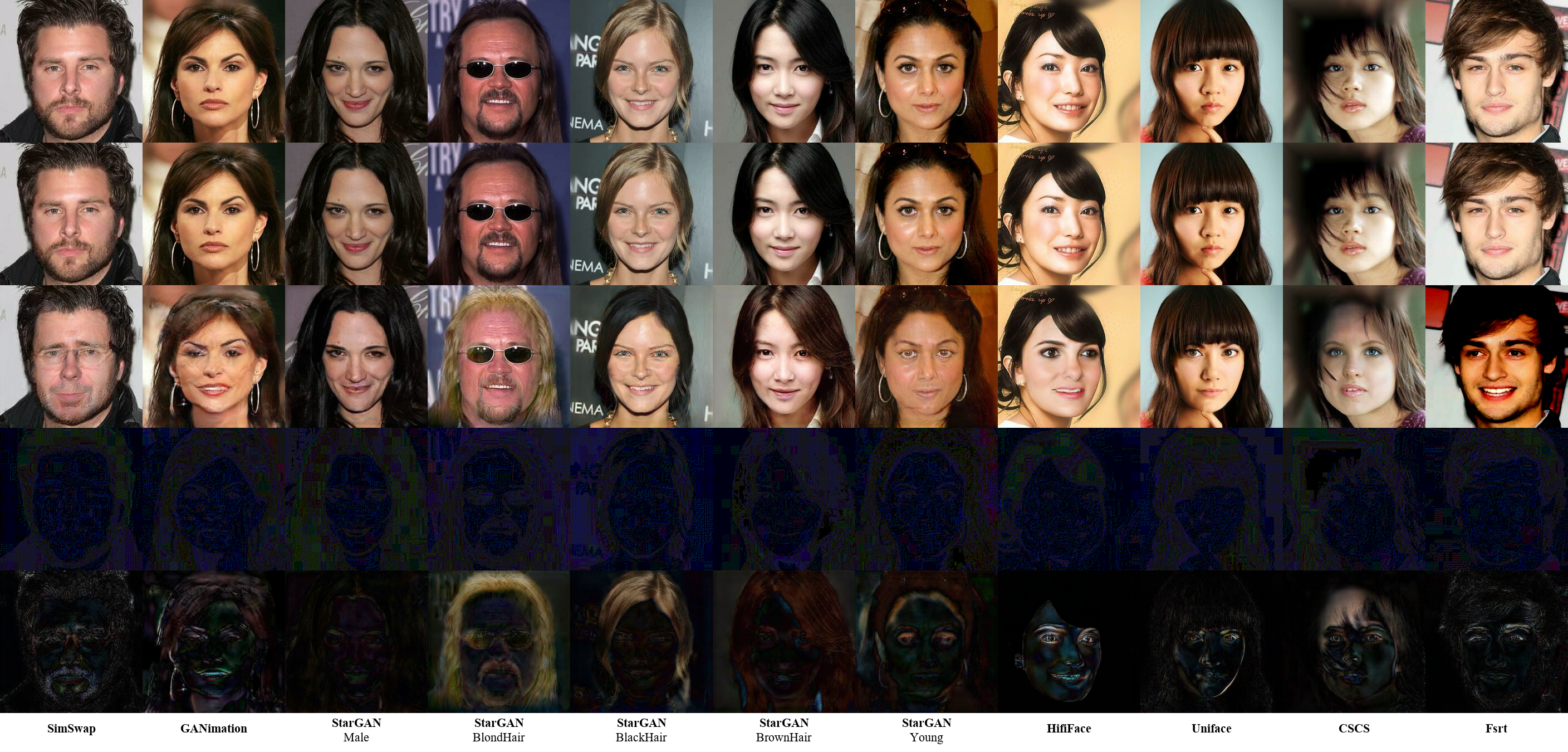} 
    
    \caption{More malicious distortions. From top to bottom are the origin image $I_{o}$, the watermarked image $I_{w}$, the noised image $I_{n}$, the residual signal $|I_{o} - I_{w}|$, and $|I_{n} - I_{w}|$. All images are resized to $256 \times 256$ for consistency.}
    \label{fig:6}
\end{figure*}

\textbf{Manipulation Pools.} In order to comprehensively evaluate the robustness and adaptability of the algorithm under different distortion conditions, this study constructs two types of operation pools: the Common Pool and Malicious Pool. 
The Common Pool includes a variety of common image distortion operations, such as \textit{\{JpegTest, Resize, GaussianBlur, MedianBlur, Dropout, SaltPepper, GaussianNoise\} }.
These operations are used in the training and testing stages to simulate the general distortions that may occur in real-world scenarios, while ensuring that the facial content of the image remains unchanged.
The Malicious Pool includes two common types of deepfake technologies: face swap and face replay. 
Specifically, for face swapping, we employ SimSwap\cite{swap1}, Uniface\cite{uniface}, CSCS\cite{cscs}, HifiFace\cite{hififace}, and FSRT\cite{fsrt}. These methods span various architectural paradigms, including identity injection (SimSwap), latent space alignment based on StyleGAN (Uniface), explicit pseudo-input supervision (CSCS), 3D-aware face modeling (HiFiFace), and transformer-based disentanglement of appearance, pose, and expression (FSRT). For face reenactment, we adopt GANimation\cite{replace8} and StarGAN\cite{replace3}, which manipulate facial expressions and attributes through expression-guided deformation and multi-domain attribute control, respectively. 
In the face swap task, the target identity is randomly selected from the CelebA verification set. For reenactment models, the target expressions and attributes are randomly sampled from a predefined attribute set: \textit{Male, Young, BlackHair, BlondHair, BrownHair}. 
All forged images are generated using the publicly released pre-trained models to ensure the high credibility and diversity of deepfake images.
\begin{table}[!t]
\centering
\footnotesize
\renewcommand{\arraystretch}{1.2}
\setlength{\tabcolsep}{3pt} 
\caption{Evaluation of Watermarked Image Quality on CelebA-HQ Using PSNR and SSIM Across Different Models, Image Resolutions, and Message Lengths
}
\label{tab:1}
\tiny 
\setlength{\tabcolsep}{5pt} 
\renewcommand{\arraystretch}{1.2} 
\resizebox{\columnwidth}{!}{%
\begin{tabular}{@{}lcccc@{}}
\toprule
Model & \makecell{Image\\Size} & \makecell{Message\\Length} & PSNR ↑ & SSIM ↑\\ \midrule
MBRS\cite{mbrs}           & $128\times128$ & 30  & 33.0456         & 0.8106                  \\
CIN\cite{cin}            & $128\times128$ & 30  & 42.4135         & 0.9629          \\
PiMoG\cite{pimog}          & $128 \times 128$ & 30  & 37.7271         & 0.9470                  \\
SepMark\cite{sepmark}        & $128 \times 128$ & 30  & 38.5112         & 0.9588                   \\
ARWGAN\cite{arwgan}        & $128 \times 128$ & 30  & 38.5746        & 0.9733                 \\
EditGuard \cite{editguard}        & $128 \times 128$ & 30  & 38.1664         & 0.9796                  \\
LampMark\cite{lampmark}        & $128 \times 128$ & 30  & 44.6251         & 0.9921                  \\
\textbf{Ours}  & $128 \times 128$ & 30  & \textbf{47.5121} & \textbf{0.9957}   \\ \midrule
SepMark\cite{sepmark}        & $256 \times 256$ & 128 & 38.5646         & 0.9328                  \\
ARWGAN\cite{arwgan}      & $256 \times 256$ & 128 & 31.3319         & 0.9339                 \\
FaceSigns\cite{facesigns}      & $256 \times 256$ & 128 & 32.3319         & 0.9211                 \\
EditGuard\cite{editguard}      & $256 \times 256$ & 128 & 37.8254        & 0.9512                 \\
LampMark\cite{lampmark}      & $256 \times 256$ & 128 & 42.7825        & 0.9869                 \\
\textbf{Ours}  & $256 \times 256$ & 128 & \textbf{43.2530} & \textbf{0.9909}   \\ \bottomrule
\end{tabular}%
}
\end{table}

\begin{figure}
    \centering
    \includegraphics[width=\linewidth]{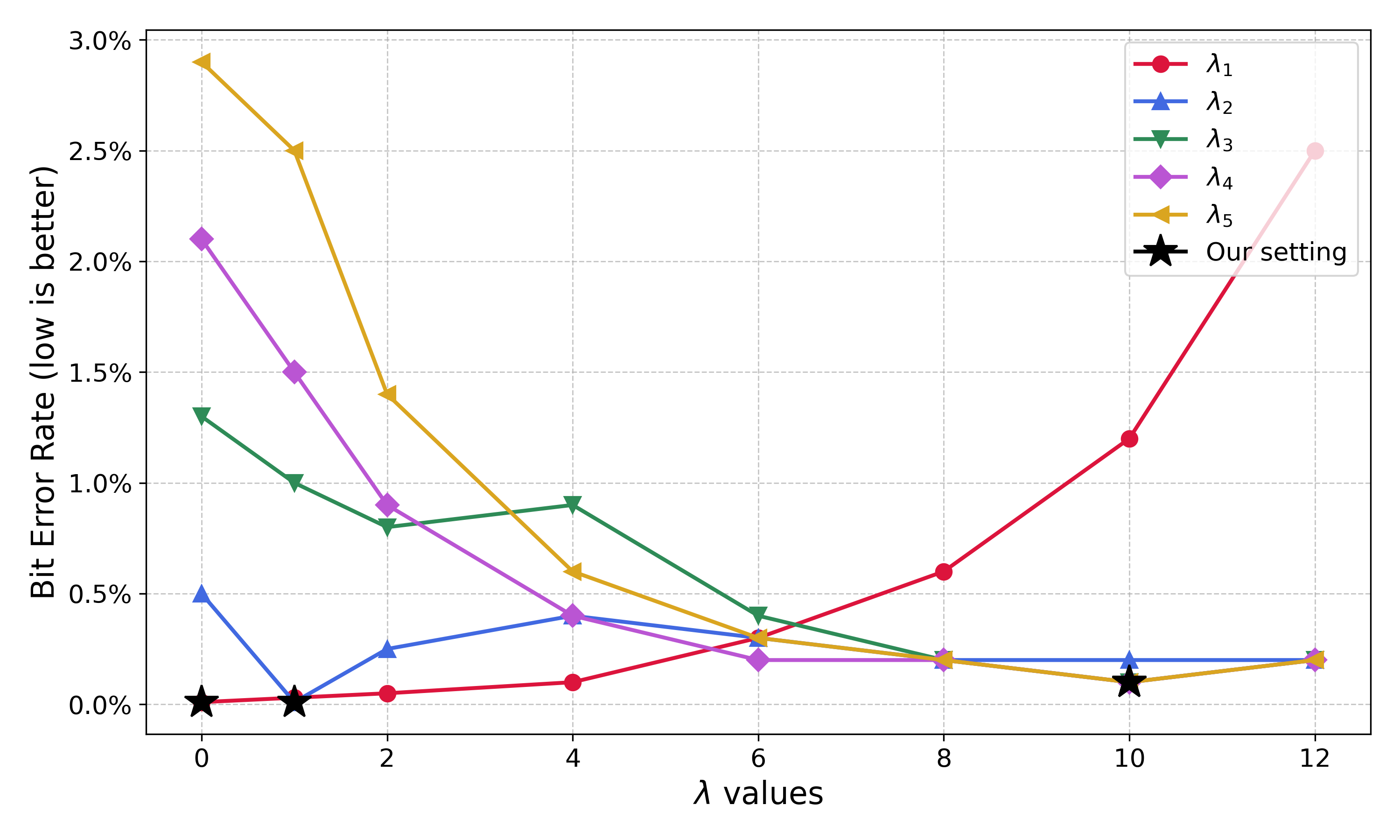}
    \caption{Hyperparameter sensitivity analysis of loss Weights ($\lambda_1$–$\lambda_5$) on bit error rate.}
    \label{fig:8}
    \vspace{-0.5cm}
\end{figure}

\textbf{Parameters.} This research is based on PyTorch and runs on an NVIDIA RTX 4090 GPU, with the random seed fixed to 42 to ensure reproducibility.
The Adam optimizer\cite{adam} is initialized with a learning rate of $5\times10^{-4}$, which is decayed stepwise to $5\times10^{-5}$ and $1\times10^{-6}$ at the 3rd and 7th epochs, respectively, to stabilize training. The model is trained with a batch size of 32 for 10 epochs, as preliminary experiments show that performance converges around the 8th–10th epoch, and longer training yields limited improvements while increasing the risk of overfitting. The embedding strength is fixed at $\alpha=0.04$.
The total loss function is composed of multiple loss terms, which are combined by weighting to balance the influence of each loss on model optimization. 
As shown in Formula (7), the weight parameters for each loss term during training are set to $\lambda_1$, $\lambda_2$, $\lambda_3$, $\lambda_4$, and $\lambda_5$ as 0.01, 1, 10, 10, and 10, respectively. To verify the appropriateness of these weights, we conducted a hyperparameter sensitivity analysis, as illustrated in Fig. \ref{fig:8}. 
This figure shows the impact of varying each $\lambda$ on the bit error rate. The black star markers denote the adopted settings, each yielding near-optimal performance for its corresponding loss component.
These values are determined based on preliminary experimental results and empirical observations, reflecting the relative contribution of each task objective to the overall optimization. The goal is to achieve a well-balanced trade-off and stable convergence among embedding quality, detection performance, and structural consistency.

\begin{table*}[t]
\centering
\footnotesize
\renewcommand{\arraystretch}{1.2}
\setlength{\tabcolsep}{2pt}
\caption{Robustness evaluation of watermark recovery on CelebA-HQ under common and deepfake distortions at 128$\times$128 resolution, measured by bit error rate (BER).}
\label{tab:2}

\begin{adjustbox}{width=\textwidth}  
\begin{tabular}{ccccccc|cc|cc}
\specialrule{0.15em}{0pt}{0pt}
\multirow{2}{*}{\bf Distortion}
& \multirow{2}{*}{MBRS\cite{mbrs}}
& \multirow{2}{*}{CIN\cite{cin}}
& \multirow{2}{*}{PIMoG\cite{pimog}}
& \multirow{2}{*}{ARWGAN\cite{arwgan}}
& \multirow{2}{*}{EditGuard\cite{editguard}}
& \multirow{2}{*}{LampMark\cite{lampmark}}
& \multicolumn{2}{c|}{SepMark\cite{sepmark}}
& \multicolumn{2}{c}{\textbf{Ours}} \\
\cline{8-11}
&&&&&&& Tracer & Detector & Tracer & Detector \\
\hline
JpegTest & 0.2597\% & 2.4317\% & 18.4971\% & 6.7001\% & 1.4152\% & 0.0000\% & 0.2136\% & 0.2172\% & 0.0000\% & 0.0000\% \\
Resize & 0.0000\% & 0.0000\% & 0.0612\% & 0.0121\% & 0.0017\% & 0.0000\% & 0.0059\% & 0.0212\% & 0.0000\% & 0.0000\% \\
GaussianBlur & 0.0000\% & 19.3915\% & 0.2312\% & 4.2970\% & 1.6027\% & 0.2104\% & 0.0024\% & 0.0035\% & 0.0021\% & 0.0032\% \\
MedianBlur & 0.0000\% & 0.0325\% & 0.0992\% & 3.0027\% & 1.6212\% & 0.1023\% & 0.0012\% & 0.0012\% & 0.0011\% & 0.0019\% \\
Dropout & 0.0000\% & 0.0000\% & 0.4713\% & 0.6905\% & 0.9257\% & 0.0000\% & 1.9125\% & 0.0000\% & 0.0000\% & 0.0000\% \\
SaltPepper & 0.0000\% & 0.0364\% & 2.1571\% & 0.0000\% & 0.0925\% & 5.1365\% & 0.0413\% & 0.0106\% & 0.0187\% & 0.0211\% \\
GaussianNoise & 0.0000\% & 0.0000\% & 11.4598\% & 6.3725\% & 1.6772\% & 0.2561\% & 0.7460\% & 0.8735\% & 0.0253\% & 0.0361\% \\
\hline
Average & 0.0371\% & 3.1274\% & 4.7110\% & 3.0107\% & 1.0480\% & 0.8150\% & 0.4176\% & 0.1610\% & \textbf{0.0067\%} & \textbf{0.0089\%} \\
\specialrule{0.15em}{0pt}{0pt}
FSRT & 2.4113\% & 3.2072\% & 5.0021\% & 4.3410\% & 5.7701\% & 2.9307\% & 0.7814\% & 48.3815\% & 0.0785\% & 46.7852\% \\
SimSwap & 24.6044\% & 40.0828\% & 7.6547\% & 46.6012\% & 45.3202\% & 0.1975\% & 20.0295\% & 50.8829\% & 0.1637\% & 46.7815\% \\
CSCS & 10.1000\% & 0.2904\% & 0.2512\% & 6.2905\% & 0.9901\% & 2.3015\% & 0.6801\% & 48.7418\% & 0.0091\% & 46.7879\% \\
HifiFace & 2.7251\% & 7.4510\% & 1.0211\% & 14.2084\% & 6.7104\% & 2.9713\% & 0.9904\% & 48.2511\% & 0.0081\% & 46.7872\% \\
Uniface & 0.4802\% & 11.8027\% & 2.7854\% & 26.2814\% & 9.1702\% & 6.2801\% & 0.3430\% & 48.9112\% & 0.0052\% & 46.7871\% \\
GANimation & 0.0000\% & 0.0000\% & 0.4501\% & 0.5201\% & 0.7152\% & 1.0254\% & 0.0000\% & 36.7938\% & 0.0000\% & 46.7873\% \\
StarGAN (Male) & 15.4416\% & 56.9345\% & 9.3444\% & 36.7825\% & 7.6204\% & 17.5317\% & 0.1125\% & 52.7725\% & 0.0065\% & 46.7869\% \\
StarGAN (Young) & 17.3462\% & 61.0875\% & 8.8415\% & 36.4812\% & 8.1745\% & 17.1141\% & 0.1560\% & 52.7907\% & 0.0071\% & 46.7875\% \\
StarGAN (BlackHair) & 19.1083\% & 58.7913\% & 9.1392\% & 36.2510\% & 7.3674\% & 18.5142\% & 0.1346\% & 48.2134\% & 0.0032\% & 46.7876\% \\
StarGAN (BlondHair) & 18.1408\% & 54.1533\% & 11.3514\% & 36.1451\% & 9.2541\% & 17.3256\% & 0.1172\% & 48.4184\% & 0.0045\% & 46.7876\% \\
StarGAN (BrownHair) & 17.6082\% & 69.5127\% & 8.9910\% & 35.9919\% & 8.1254\% & 17.4425\% & 0.0997\% & 48.1250\% & 0.0064\% & 46.7872\% \\
\hline
Average & 11.6332\% & 33.0284\% & 5.8938\% & 25.4449\% & 9.9289\% & 9.4213\% & 2.1313\% & 48.3893\% & \textbf{0.0266\%} & 46.7848\% \\
\specialrule{0.15em}{0pt}{0pt}
\end{tabular}
\end{adjustbox}
\end{table*}

\textbf{Baselines. }To verify the performance of the proposed method, we selected several representative watermarking and proactive watermarking frameworks as baselines. Although this work focuses on proactive watermarking, due to the limited existing methods and scarce open-source codes, robust watermarking methods are also included to evaluate stability and extraction performance.
In the watermarking domain, we select five robust methods: MBRS \cite{mbrs}, CIN \cite{cin} and PIMOG \cite{pimog} (all only for 128×128), as well as ARWGAN \cite{arwgan}, and EditGuard \cite{editguard}.
For proactive watermarking, we include FaceSigns \cite{facesigns} (only for 256×256), SepMark \cite{sepmark}, and LampMark \cite{lampmark}.
Due to differences in resolution compatibility, each method is reported only in the tables corresponding to its applicable settings.
All the other baseline methods use publicly available pre-trained model weights to ensure fairness. 
\enlargethispage*{0pt}

\subsection{Visual Quality}
To evaluate the visual quality of the image after watermark embedding, this paper adopts two indicators: average peak signal-to-noise ratio (PSNR) and structural similarity index measure (SSIM), which respectively quantify the changes in image signal distortion and structural fidelity, fully reflecting the impact of embedding on the image. 
TABLE \ref{tab:1} quantitatively verifies that our method has the best visual quality in all baselines and shows better invisibility.
Experimental results show that our method achieves the highest PSNR and SSIM values at resolutions of 128×128 and 256×256.
This improvement is primarily attributed to SC-GNN and attention-enhanced embedding.
In addition, Fig.\ref{fig:5} and \ref{fig:6} show the subjective visual contrast results at different resolutions, indicating that the watermark image is still highly consistent with the original image after embedding, and it is almost difficult to detect the visual difference.
Combining the objective evaluation results in TABLE \ref{tab:1} with the subjective visual comparisons in Fig. \ref{fig:5} and \ref{fig:6}, it is evident that the proposed method achieves superior imperceptibility in both quantitative metrics and human visual perception.

\subsection{Robustness Test}
The Bit Error Rate (BER) is selected as the main evaluation index in the robustness test. The BER calculates the ratio of error bits to the total length
of the watermark, serving as a direct measure of recovery
accuracy under various distortion scenarios.
\begin{equation}
\text{BER}(W_{\text{rec}}, W_o) = \frac{\sum_{i=0}^{l-1} \left| W_{\text{rec}}^i - W_o^i \right|}{l}
\label{eq:ber}
\end{equation}
where \( W_o \) is the original watermark, \( W_{\text{rec}} \) represents the recovered watermark, and \( l \) denotes the length of the binary watermark sequence. 
In any case, the BER value of Tracer should be close to 0\%, and the BER value of the semi-robust Detector should be close to 50\% when it encounters malicious distortion. In the case of common distortion, the BER value should be close to 0\%. 
This 50\% threshold is not a strict rule but a practical choice, as our method naturally exhibits a clear gap between the highly robust Tracer and the semi-robust Detector. Such contrast makes the detection process intuitive and effective without requiring parameter tuning. A similar threshold-based design is also adopted in SepMark\cite{sepmark}.
Tables \ref{tab:2} and  \ref{tab:3} provide a quantitative comparison of BER performance across different methods under both common distortions and deepfake forgeries, evaluated at 128×128 and 256×256 resolutions, respectively.
Experimental results show that our method achieves the lowest BER in all test scenarios, showing excellent robustness.
Under ordinary distortion, the BER of Tracer and Detector are both close to 0\%, which ensures that the watermark can be recovered with high precision under all kinds of interference. 
\begin{table*}[t]
\centering
\small
\renewcommand{\arraystretch}{1.2}
\caption{Robustness evaluation of watermark recovery on CelebA-HQ under common and deepfake distortions at 256$\times$256 resolution, measured by bit error rate (BER).}
\label{tab:3}

\begin{adjustbox}{width=\textwidth}
\begin{tabular}{ccccc|cc|cc}
\specialrule{0.15em}{0pt}{0pt}
\multirow{2}{*}{\bf Distortion}
& \multirow{2}{*}{ARWGAN\cite{arwgan}}
& \multirow{2}{*}{FaceSigns\cite{facesigns}}
& \multirow{2}{*}{EditGuard\cite{editguard}}
& \multirow{2}{*}{LampMark\cite{lampmark}}
& \multicolumn{2}{c|}{SepMark\cite{sepmark}}
& \multicolumn{2}{c}{\textbf{Ours}} \\
\cline{6-9}
&&&&& Tracer & Detector & Tracer & Detector \\
\hline
JpegTest & 6.0034\% & 0.6298\% & 3.5801\% & 0.0000\% & 0.0071\% & 0.0071\% & 0.0000\% & 0.0000\% \\
Resize & 0.2501\% & 1.3654\% & 0.2815\% & 1.5809\% & 0.0000\% & 0.0000\% & 0.0000\% & 0.0000\% \\
GaussianBlur & 7.6214\% & 0.1971\% & 3.3701\% & 0.3151\% & 0.0000\% & 0.0298\% & 0.0000\% & 0.0056\% \\
MedianBlur & 3.1214\% & 0.0997\% & 3.8504\% & 0.1825\% & 0.0000\% & 0.0000\% & 0.0000\% & 0.0000\% \\
Dropout & 0.7285\% & 18.4514\% & 0.9752\% & 2.4312\% & 0.0061\% & 0.0000\% & 0.0019\% & 0.0027\% \\
SaltPepper & 0.5011\% & 11.4701\% & 0.1210\% & 5.2514\% & 0.0011\% & 0.0009\% & 0.0096\% & 0.0045\% \\
GaussianNoise & 6.2471\% & 8.1637\% & 1.0274\% & 0.8254\% & 0.0525\% & 0.0682\% & 0.0045\% & 0.0045\% \\
\hline
Average &  3.4961\% & 5.7682\% & 1.8865\% & 1.5124\% & 0.0095\% & 0.0151\% & \textbf{0.0023\%} & \textbf{0.0025\%} \\
\specialrule{0.15em}{0pt}{0pt}
FSRT & 4.9701\% & 49.1201\% & 37.1724\% & 3.2541\% & 9.2103\% & 45.2127\% & 0.0091\% & 44.5118\% \\
SimSwap & 41.5726\% & 49.3975\% & 47.5625\% & 0.7125\% & 7.9125\% & 45.9127\% & 0.0096\% & 44.5119\% \\
CSCS & 14.8215\% & 49.6731\% & 1.6173\% & 5.1402\% & 2.3514\% & 45.9117\% & 0.0081\% & 44.5153\% \\
HifiFace & 11.2041\% & 49.7431\% & 45.0121\% & 9.1074\% & 14.0813\% & 45.8724\% & 0.0076\% & 44.5125\% \\
Uniface & 21.8204\% & 49.6547\% & 49.4123\% & 7.0297\% & 18.9325\% & 45.9117\% & 0.0086\% & 44.5157\% \\
GANimation & 4.3051\% & 45.3601\% & 7.5911\% & 0.0165\% & 0.0020\% & 43.8325\% & 0.0000\% & 44.5131\% \\
StarGAN (Male) & 32.5146\% & 50.5931\% & 2.1241\% & 16.1040\% & 0.0073\% & 45.4375\% & 0.0013\% & 44.5123\% \\
StarGAN (Young) & 32.1901\% & 50.4131\% & 3.5414\% & 16.2167\% & 0.0030\% & 45.3125\% & 0.0009\% & 44.5127\% \\
StarGAN (BlackHair) & 31.9904\% & 50.2376\% & 2.9812\% & 15.4211\% & 0.0016\% & 45.3644\% & 0.0032\% & 44.5178\% \\
StarGAN (BlondHair) & 32.0541\% & 50.9129\% & 2.8574\% & 16.0487\% & 0.0021\% & 44.6956\% & 0.0011\% & 44.5131\% \\
StarGAN (BrownHair) & 32.5706\% & 50.7404\% & 3.0145\% & 16.4575\% & 0.0037\% & 45.6759\% & 0.0021\% & 44.5179\% \\
\hline
Average & 23.6376\% & 49.6223\% & 18.4442\% & 9.5917\% &  4.7734\% & 46.2854\% & \textbf{0.0047\%} & 44.5066\% \\
\specialrule{0.15em}{0pt}{0pt}
\end{tabular}
\end{adjustbox}
\end{table*}

However, the BER of the contrast method is obviously higher under various common distortion conditions, which indicates that its robustness is weak.
Under the deepfake forgery, the BER of Tracer is still close to 0\%, while the BER of semi-robust Detector is maintained close to 50\%, which effectively distinguishes the true and forged contents. 
This advantage mainly stems from our combined design of frequency-domain embedding and structural modeling. Existing methods often perform poorly when facing semantic-level transformations, making watermarks more vulnerable. For example, SimSwap reconstructs the entire face, overwriting original pixels and disrupting spatial continuity, which leads to high BER for shallow feature based methods. In contrast, our approach enhances watermark redundancy through DT-CWT in directional sub-bands and leverages SC-GNN to model structural relationships between regions, enabling robust watermark recovery even under significant appearance and semantic changes.
Although our proposed semi-robust Detector exhibits a slightly lower BER than SepMark under deepfake attacks, the primary objective of the Detector is to distinguish between authentic and forged content. As long as a significant BER gap exists between common distortions and deepfake forgeries, effective detection can be achieved. Our method yields a substantially lower BER under common distortions, thereby enlarging the separation margin between real and fake samples and significantly enhancing discriminability.
However, the existing methods have large fluctuations in BER and lack of detection reliability.
\begin{table}[t]
\centering
\footnotesize
\renewcommand{\arraystretch}{1.25}
\setlength{\tabcolsep}{5pt}
\caption{Cross-dataset robustness and visual quality on LFW, FaceForensics++, DFDC, and Celeb-DF v2. Visual quality is reported with PSNR/SSIM, and robustness with BER (\%). $\Delta$ denotes the absolute BER difference compared with CelebA-HQ.}
\label{tab:4}
\resizebox{\linewidth}{!}{ 
\begin{tabular}{c|c|c|c|c}
\specialrule{0.1em}{0pt}{0pt}
\multicolumn{5}{c}{128 $\times$ 128} \\
\hline
Metric / Distortion & LFW & FF++ & DFDC & Celeb-DF v2 \\
\hline
PSNR ↑  & 46.27 & 47.65 & 47.65 & 47.67 \\
SSIM ↑  & 0.9948 & 0.9957 & 0.9945 & 0.9957 \\
\hline
Average Common   & 0.0101\% & 0.0098\% & 0.0059\% & 0.0076\% \\
$\Delta$         & $\uparrow$\textbf{0.0034\%} & $\uparrow$\textbf{0.0031\%} & $\downarrow$\textbf{0.0008\%} & $\uparrow$\textbf{0.0009\%} \\
Average Malicious& 0.0389\% & 0.0278\% & 0.0225\% & 0.0246\% \\
$\Delta$         & $\uparrow$\textbf{0.0123\%} & $\uparrow$\textbf{0.0012\%} & $\downarrow$\textbf{0.0041\%} & $\downarrow$\textbf{0.0020\%} \\
\specialrule{0.1em}{0pt}{0pt}
\multicolumn{5}{c}{256 $\times$ 256} \\
\hline
Metric / Distortion & LFW & FF++ & DFDC & Celeb-DF v2 \\
\hline
PSNR ↑  & 43.11 & 43.41 & 43.37 & 43.39 \\
SSIM ↑  & 0.9908 & 0.9895 & 0.9897 & 0.9894 \\
\hline
Average Common   & 0.0036\% & 0.0017\% & 0.0028\% & 0.0038\% \\
$\Delta$         & $\uparrow$\textbf{0.0013\%} & $\downarrow$\textbf{0.0006\%} & $\uparrow$\textbf{0.0005\%} & $\uparrow$\textbf{0.0015\%} \\
Average Malicious& 0.0035\% & 0.0058\% & 0.0045\% & 0.0061\% \\
$\Delta$         & $\downarrow$\textbf{0.0012\%} & $\uparrow$\textbf{0.0011\%} & $\downarrow$\textbf{0.0002\%} & $\uparrow$\textbf{0.0014\%} \\
\specialrule{0.1em}{0pt}{0pt}
\end{tabular}
}
\vspace{-0.3cm}
\end{table}

\begin{table*}[htbp]

\begin{minipage}{.48\linewidth}
\caption{ablation results for psnr, ssim, and tracer’s bit error rate (\%), where ‘×’ denotes non-usage and ‘√’ denotes usage.
}
\label{tab:5}
\setlength{\tabcolsep}{1.5pt}
\renewcommand{\arraystretch}{1.4}

\resizebox{\linewidth}{!}{
\begin{tabular}{cccccc}\specialrule{0.15em}{0pt}{0pt}
& Case1 & Case2 & Case3 & Case4 & Proposed \\
\cline { 2 - 6 } Joint Sub-bands & $\times$ & $\checkmark$ & $\times$ & $\checkmark$ & $\checkmark$ \\
Attention module & $\checkmark$ & $\times$ & $\checkmark$ & $\checkmark$ & $\checkmark$ \\
frequency domain & $\checkmark$ & $\checkmark$ & $\times$ & $\checkmark$ & $\checkmark$ \\
SC-GNN & $\checkmark$ & $\checkmark$ & $\checkmark$ & $\times$ & $\checkmark$ \\
\hline PSNR $\uparrow$ & 45.1478 & 43.6807 & 48.0785 & 42.9015 & 47.5121\\
SSIM $\uparrow$ & 0.9956 & 0.9823 & 0.9955 & 0.9417 & 0.9957 \\ \hline
JpegTest & 0.0000\% & 0.0000\% & 0.0298\% & 0.0000\% & 0.0000\% \\
Resize & 0.0000\% & 0.0000\% & 0.0157\% & 0.0000\% & 0.0000\% \\
GaussianBlur & 0.0212\% & 0.0027\% & 0.0229\% & 0.0512\% & 0.0021\% \\
MedianBlur & 0.0023\% & 0.0015\%& 0.0071\%& 0.0084\% & 0.0011\% \\
Dropout & 0.0009\% & 0.0000\% & 0.0000\% & 0.0000\% & 0.0000\% \\
SaltPepper & 0.0043\% & 0.0043\% & 0.0408\% & 0.1251\% & 0.0187\% \\ 
GaussianNoise & 0.0041\% & 0.0195\% & 0.5372\% & 0.0808\% & 0.0253\% \\
\hline
Average &0.0047\% &0.0040\% &0.0934\% &0.0379\% &0.0067\% \\ \specialrule{0.15em}{0pt}{0pt}
FSRT & 0.0825\% & 0.0232\% & 0.1984\% & 0.0813\% & 0.0785\% \\
SimSwap & 0.2164\% & 0.2139\% & 0.5991\% & 0.1905\% & 0.1637\% \\
CSCS & 0.0098\% & 0.0056\% & 0.2025\% & 0.1221\% & 0.0091\% \\
HifiFace & 0.0074\% & 0.0088\% & 0.1109\% & 0.0113\% & 0.0081\% \\
Uniface & 0.0087\% & 0.0047\% & 0.1071\% & 0.0623\% & 0.0052\% \\
GANimation & 0.0015\% & 0.0000\% & 0.0138\% & 0.0013\% & 0.0000\% \\
StarGAN (Male) & 0.0204\% & 0.0075\% & 0.1245\% & 0.0169\% & 0.0065\% \\
StarGAN (Young) & 0.0112\% & 0.0081\% & 0.1278\% & 0.0114\% & 0.0071\% \\
StarGAN (BlackHair) & 0.0103\% & 0.0094\% & 0.1138\% & 0.0121\% & 0.0032\% \\
StarGAN (BlondHair) & 0.0109\% & 0.0089\% & 0.1345\% & 0.0153\% & 0.0045\% \\
StarGAN (BrownHair) & 0.0176\% & 0.0078\% & 0.1374\% & 0.0187\% & 0.0064\% \\ \hline
Average &0.0389\% &0.0289\% &0.1759\% &0.0532\% &0.0266\% \\ 
\specialrule{0.15em}{0pt}{0pt}
\end{tabular}}
\end{minipage}
\quad
\begin{minipage}{.48\linewidth}
\vspace*{-0.8cm}
\caption{ ablation results for psnr, ssim, and detector’s bit error rate (\%), where ‘×’ denotes non-usage and ‘√’ denotes usage.
}
\label{tab:6}
\setlength{\tabcolsep}{1.5pt}
\renewcommand{\arraystretch}{1.4}

\resizebox{\linewidth}{!}{
\begin{tabular}{cccccc}\specialrule{0.15em}{0pt}{0pt}
& Case1 & Case2 & Case3 & Case4 & Proposed \\
\cline { 2 - 6 } Joint Sub-bands & $\times$ & $\checkmark$ & $\times$ & $\checkmark$ & $\checkmark$ \\
Attention module & $\checkmark$ & $\times$ & $\checkmark$ & $\checkmark$ & $\checkmark$ \\
frequency domain & $\checkmark$ & $\checkmark$ & $\times$ & $\checkmark$ & $\checkmark$ \\
SC-GNN & $\checkmark$ & $\checkmark$ & $\checkmark$ & $\times$ & $\checkmark$ \\
\hline PSNR $\uparrow$ & 45.1478 & 43.6807 & 48.0758 & 42.9015 & 47.5121 \\
SSIM $\uparrow$ & 0.9956 & 0.9823 & 0.9955 & 0.9417 & 0.9957 \\ \hline
JpegTest & 0.0000\% & 0.0000\% & 0.0174\% & 0.0000\% & 0.0000\% \\
Resize & 0.0000\% & 0.0000\% & 0.0324\% & 0.0000\% & 0.0000\% \\
GaussianBlur & 0.0028\% & 0.0037\% & 0.0064\% & 0.0074\% & 0.0032\% \\
MedianBlur & 0.0051\% & 0.0044\% & 0.0061\% & 0.0027\% & 0.0019\% \\
Dropout & 0.0000\% & 0.0000\% & 0.0000\% & 0.0000\% & 0.0000\% \\
SaltPepper & 0.0277\% & 0.0298\% & 0.0308\% & 0.0081\% & 0.0211\% \\
GaussianNoise & 0.0149\% & 0.0121\% & 0.7391\% & 0.0209\% & 0.0361\% \\
\hline
Average &0.0072\% &0.0071\% &0.1189\% &0.0056\% &0.0089\% \\ \specialrule{0.15em}{0pt}{0pt}
FSRT & 45.4874\% & 46.3172\% & 46.7846\% & 46.1524\% & 46.7852\% \\
SimSwap & 45.4821\% & 46.3141\% & 46.7871\% & 46.1574\% &46.7815\% \\
CSCS & 45.4832\% & 46.3164\% & 46.7873\% & 46.1524\% & 46.7879\% \\
HifiFace & 45.4852\% & 46.3119\% & 46.7877\% & 46.1499\% & 46.7872\% \\
Uniface & 45.4840\% & 46.3177\% & 46.7857\% & 46.1577\% & 46.7871\% \\
GANimation & 45.4851\% & 46.3134\% & 46.7856\% & 46.1525\% & 46.7873\%\\
StarGAN (Male) & 45.4834\% & 46.3122\% & 46.7836\% & 46.1531\% & 46.7869\% \\
StarGAN (Young) & 45.4822\% & 46.3188\% & 46.7856\% & 46.1510\% & 46.7875\% \\
StarGAN (BlackHair) & 45.4831\% & 46.3139\% & 46.7862\% & 46.1584\% & 46.7876\% \\
StarGAN (BlondHair) & 45.4816\% & 46.3147\% & 46.7843\%& 46.1545\% & 46.7876\% \\
StarGAN (BrownHair) & 45.4843\% & 46.3147\% & 46.7851\% & 46.1547\% & 46.7872\% \\ \hline
Average &45.4738\% &46.3162\% &46.7857\% &46.1586\% &46.7848\% \\ 
\specialrule{0.15em}{0pt}{0pt}
\end{tabular}}
\end{minipage}
\end{table*}

To evaluate the cross-dataset generalization ability, TABLE~\ref{tab:4} reports the BER performance of our method on four datasets (LFW\cite{LFW_tcsvt}, FaceForensics++\cite{FF++}, DFDC\cite{DFDC}, and Celeb-DF v2\cite{celeb-df}) under both common and malicious distortions at two resolutions. The row denoted as $\Delta$ represents the absolute BER difference compared with the original training dataset (CelebA-HQ), with arrows indicating the direction of change---$\uparrow$ for increased BER and $\downarrow$ for decreased BER. Although the datasets differ significantly in distribution, our method consistently maintains low BER in tracing across all datasets and distortion types, demonstrating strong robustness and reliable transferability to unseen data.

This robustness is mainly due to the frequency-domain embedding strategy, which plays a central role in ensuring generalization across datasets. By distributing watermark information across multiple directional sub-bands, the DT-CWT enhances redundancy and stability under various distortions and data distributions. In addition, the SC-GNN module provides structural modeling that helps the system capture cross-domain spatial consistencies, further improving robustness against semantic and appearance variations. Finally, the feature reuse mechanism is introduced to improve the flexibility and robustness of watermark embedding by optimizing the feature transfer process. 
\begin{figure}[!t]
    \centering
    \includegraphics[width=\linewidth]{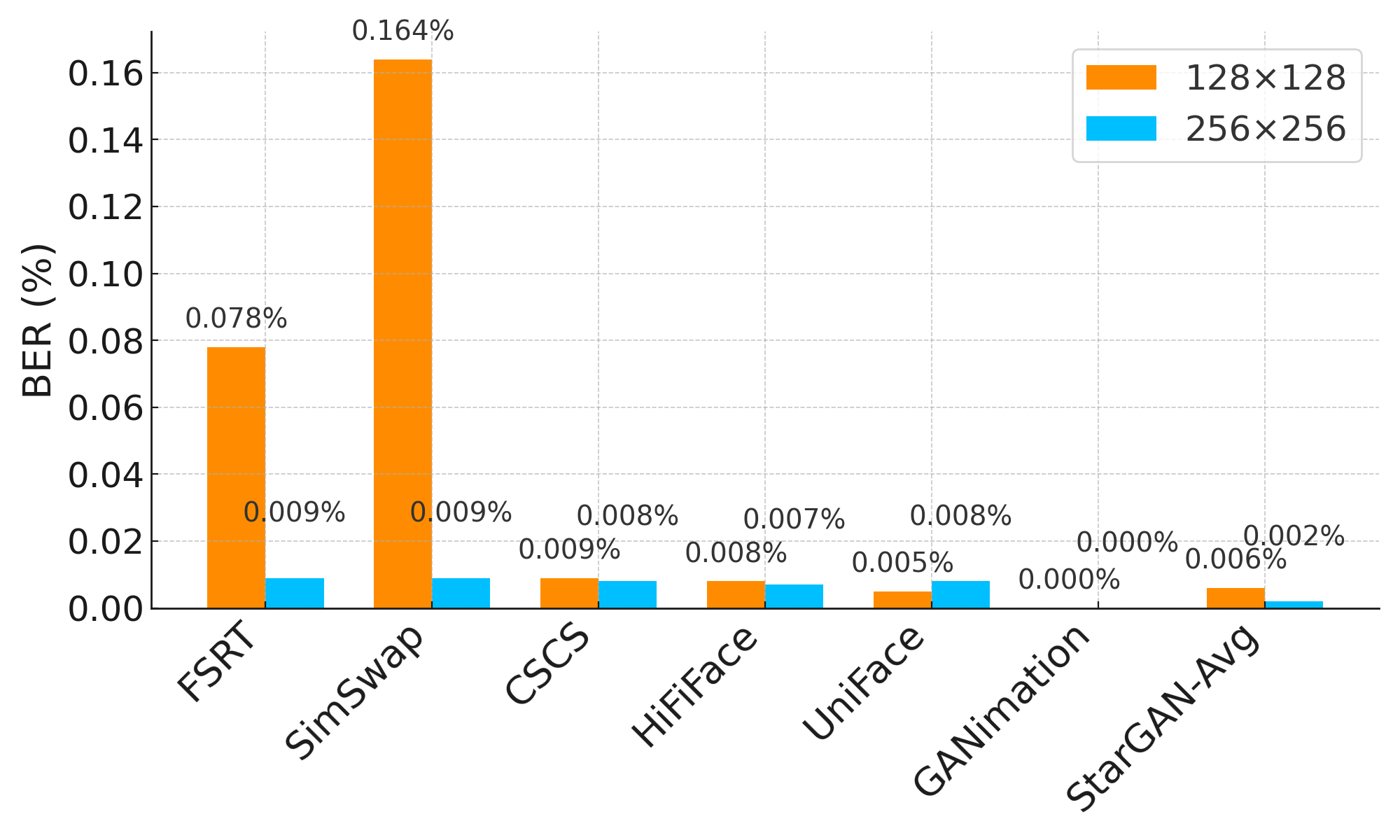}
    \caption{Sensitivity analysis of watermark robustness under diverse deepfake distortions at varying resolutions.}
    \label{fig:7}
    \vspace*{-0.5cm}
\end{figure}

\begin{table}[t]
\centering
\footnotesize
\renewcommand{\arraystretch}{1.25}
\caption{Computational complexity of WaveGuard: parameters, FLOPs, and inference time on an NVIDIA RTX 4090 GPU.}
\label{tab:7}
\resizebox{\linewidth}{!}{
\begin{tabular}{c|c|c|c}
\specialrule{0.1em}{0pt}{0pt}
\textbf{Input Resolution} & \textbf{Params (M)} & \textbf{FLOPs (G)} & \textbf{Time (ms/image)} \\
\hline
128 $\times$ 128 & 15.5 & 3.38  & 74.0 \\
256 $\times$ 256 & 15.5 & 13.5  & 70.9 \\
\specialrule{0.1em}{0pt}{0pt}
\end{tabular}
}
\vspace{-0.5cm}
\end{table}

\vspace{-0.5cm}
\subsection{Complexity Analysis}
To comprehensively evaluate the computational feasibility of WaveGuard, we report its parameter size, FLOPs, and inference speed on an NVIDIA RTX 4090 GPU, as shown in TABLE \ref{tab:7}. The model contains approximately 15.5M trainable parameters, reflecting moderate structural complexity. In terms of computational cost, WaveGuard requires about 13.5 GFLOPs to process a $256 \times 256$ image, with an average inference time of 70.9 ms per image. When the input resolution is reduced to $128 \times 128$, the computational cost drops significantly to about 3.38 GFLOPs (roughly one quarter of the $256 \times 256$ case due to the quadratic reduction in pixel count), while the inference time remains similar at around 74.0 ms per image owing to GPU parallelization. These results indicate that WaveGuard achieves a well-balanced trade-off between robustness, traceability, and computational efficiency, making its complexity acceptable for practical deepfake detection and source tracing scenarios.
\vspace{-0.3cm}
\subsection{Robustness Sensitivity Analysis}
To analyze the robustness and sensitivity of watermark embedding under different deepfake generation types and image resolutions (128×128 and 256×256), we present the evaluation results in Fig. \ref{fig:7}.
Among them, the BER of SimSwap and FSRT increases significantly at low resolution, which indicates that their operations based on semantic reconstruction and whole face replacement have the most serious damage to the watermark.
However, forgery methods such as CSCS, HifiFace and GANimation have limited changes to the image structure, so the BER is always kept at a low level. The average BER of StarGAN series is also low, indicating that the interference ability of attribute-level forgery to watermark is relatively weak.
On the whole, the robustness of 256×256 resolution is significantly better than that of 128×128, mainly due to the finer distribution and stronger redundancy of watermarks brought by higher resolution, thus having higher recovery ability under local damage.
In addition, the pre-training weights of some deepfake models (such as Uniface, HifiFace, CSCS and FSRT) only support 256×256 input, and 128×128 images need to be resize to high resolution when used. However, the interpolation process will lead to the loss of image details and edge structure, weaken the attack intensity of forgery, and indirectly improve the recoverability and robustness of the watermark.

\vspace{-0.3cm}
\subsection{Ablation Study}
In order to evaluate the contribution of Joint Sub-bands, Attention Module, Frequency Domain Embedding, and GNN in this method, we conducted ablation experiments on both Tracer and Detector. 
The experiments were conducted on images with a resolution of 128×128.
(a) Adjust the sub-bands usage of Tracer and Detector: In Tracer, watermark embedding was reduced from 1,3,4,6 sub-bands to 1,3, while in Detector, the detection was expanded from 1,3 sub-bands to 1,3,4,6. This allows us to evaluate the influence of joint sub-band information redundancy on watermark recovery and deepfake detection.
(b) Remove the attention mechanism in  the process of watermark embedding to analyze its influence on visual quality.
(c) When the DT-CWT frequency domain  embedding is removed, the watermark is directly embedded in the U channel.
(d) Removing SC-GNN to evaluate its impact on the visual quality of the watermarked image.
TABLE \ref{tab:5} and \ref{tab:6} present the results of the ablation study.

It can be found that:
(a) From Case 1, the joint sub-band significantly affects the robustness and detection ability of the watermark. 
When Tracer only extracts the watermark in 1 and 3 sub-bands, the BER increases obviously, indicating that reducing sub-bands weakens the stability of the watermark. 
However, when the Detector is extended to 1, 3, 4 and 6 sub-bands, the BER decreases, and the extra sub-bands enhance the redundancy of the watermark, making its detection more stable under common distortion and deepfake forgery.
(b) From Case 2, it can be seen that the PSNR and SSIM of the watermark image decrease after the attention mechanism is removed, which indicates that the attention mechanism has optimized the watermark embedding.
Although BER has not changed much, the invisibility and visual quality of the watermark have obviously decreased, which verifies the importance of attention mechanism in optimizing visual quality.
(c) From Case 3, it can be seen that the robustness of the watermark is significantly reduced after the frequency domain embedding is removed. 
Although PSNR and SSIM remain at a high level, BER increases significantly under common distortion and deepfake forgery. 
In addition，the perfect reconstruction, directional selectivity and approximate translation invariance of DT-CWT make frequency domain embedding more resistant to distortion and forgery, which verifies its key role in improving watermark robustness.
(d) From Case 4, it can be seen that after removing SC-GNN, the PSNR and SSIM of the watermarked image both decrease, and the BER also shows a slight increase. This indicates that SC-GNN not only plays an important role in optimizing the visual quality of the watermarked image but also provides auxiliary improvements to robustness. 
By extracting and modeling the structural features of the image, SC-GNN reduces perceptual artifacts, improves invisibility, and enhances resistance to semantic-level manipulations such as deepfake forgeries, thereby contributing to both visual quality and robustness.
Compared with more conventional perceptual (e.g., GAN discriminators), our graph-based method places greater emphasis on modeling structural relationships at the regional level. Similarly, compared with Transformer-based constraints that rely on global self-attention to capture long-range dependencies, our method can more effectively maintain spatial consistency in semantically tampered facial images, thereby resisting geometric distortions and local appearance changes, while ensuring both high visual quality and strong robustness of the watermark.

\section{CONCLUSION}
\label{sec:conclusion}
In this paper, a watermarking framework based on frequency domain embedding and  SC-GNN is proposed.
The robustness is improved by DT-CWT transform, and the detection ability of deepfake is enhanced by combining sub-band information. 
In addition, the framework leverages dense connectivity to enhance feature reuse and robustness.
SC-GNN further optimizes the visual quality of watermark embedding, reduces perceptual artifacts, and improves the invisibility of watermark by attention mechanism.
Experimental results show that this method can stably extract watermark and accurately detect tampering in various distorted and forged scenes. 
Currently, the embedding is performed in the frequency domain of the chrominance channel, which has not fully considered distortions targeting color information. In the future, we plan to explore color-aware embedding strategies to further enhance robustness under color-related perturbations and real-world forgery scenarios.
Future work will further improve its generalization ability and optimize its adaptability to complex distortion and unknown deepfake technology.

\bibliographystyle{IEEEtran}
\bibliography{reference}

\end{document}